\documentclass[final]{cvpr}

\usepackage{times}
\usepackage{epsfig}
\usepackage{graphicx}
\usepackage{amsmath}
\usepackage{amssymb}
\usepackage{amsthm}
\usepackage{booktabs}
\usepackage{array}
\usepackage{multirow}

\usepackage[pagebackref=true,breaklinks=true,colorlinks,bookmarks=false]{hyperref}

\def\cvprPaperID{2065} 
\def\confYear{CVPR 2021}

\begin{document}

\title{SUTD-TrafficQA: A Question Answering Benchmark and an Efficient Network for Video Reasoning over Traffic Events}
\author{Li Xu \qquad He Huang \qquad Jun Liu\thanks{Corresponding Author.} \\
Information Systems Technology and Design \\
Singapore University of Technology and Design \\
{\tt\small \{li\_xu, he\_huang\}@mymail.sutd.edu.sg, jun\_liu@sutd.edu.sg}}

\maketitle
\pagestyle{empty}
\thispagestyle{empty}

\begin{figure*}[htbp]
\begin{center}
    \includegraphics[width=0.8\linewidth,,trim=1.5cm 0.5cm 1.5cm 1cm,clip]{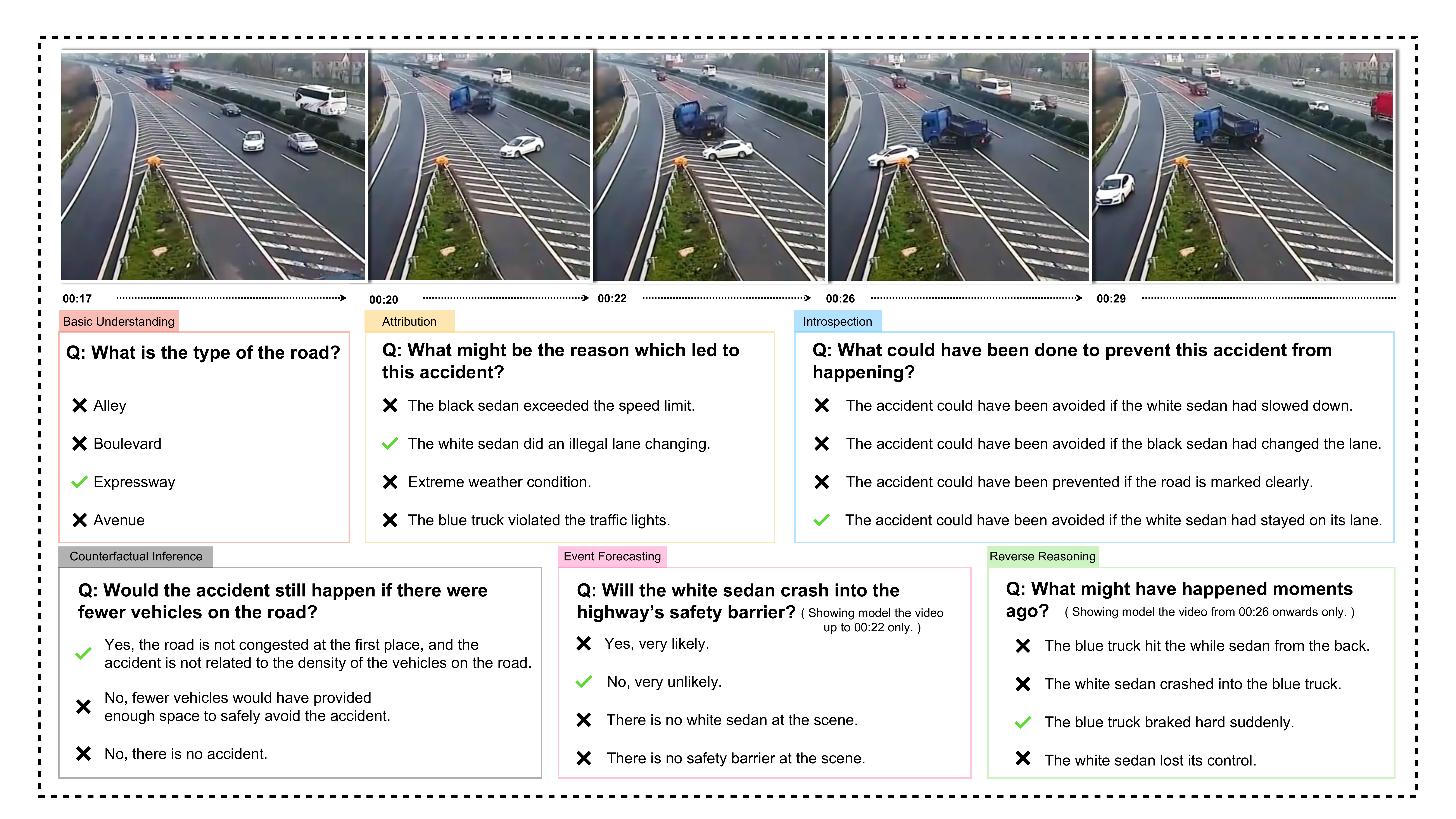}
\end{center}
\vspace{-0.6cm}
    \caption{
    An example of our SUTD-TrafficQA dataset showing that a white sedan had missed the highway exit. Hence it chose to change the lane illegally for driving towards the exit. To avoid collision, the blue truck had to brake suddenly, and then an accident occurred.
    Six reasoning tasks are designed, 
    covering a broad range of inference problems from basic understanding to complex reasoning and attribution analysis.
    To accurately answer these questions, the models need to explore the causal, logic, and spatio-temporal structures of the video event.
    }
\label{fig:example}
\vspace{-0.2cm}
\end{figure*}

\begin{abstract}
Traffic event cognition and reasoning in videos is an important task that has a wide range of applications in intelligent transportation, assisted driving, and autonomous vehicles.
In this paper, we create a novel dataset, SUTD-TrafficQA (Traffic Question Answering), which takes the form of video QA based on the collected 10,080 in-the-wild videos and annotated 62,535 QA pairs, for benchmarking the cognitive capability of 
causal inference and event understanding models in complex traffic scenarios.
Specifically, we propose 6 challenging reasoning tasks corresponding to various traffic scenarios, so as to evaluate the reasoning capability over different kinds of complex yet practical traffic events. 
Moreover, we propose \textbf{Eclipse}, a novel \textbf{E}ffi\textbf{c}ient g\textbf{li}m\textbf{pse} network via dynamic inference, in order to achieve computation-efficient and reliable video reasoning. The experiments show that our method achieves superior performance while reducing the computation cost significantly. The project page: \url{https://github.com/SUTDCV/SUTD-TrafficQA}.
%
   
\end{abstract}

\section{Introduction}

Intelligent transportation \cite{5959985} has been receiving increasing attention recently, and for the applications, such as assisted driving, violation detection, and congestion forecasting, accurate and efficient cognition and reasoning over the traffic events captured by video cameras is extremely important. As shown by previous works \cite{10.3115/1073012.1073017,35179}, well-designed datasets are often crucial for the development, adaptation and evaluation of different data-driven approaches. 
This indicates the significance of creating comprehensive and challenging benchmarks for video causal reasoning
and cognitive development of models, 
that explore the underlying causal structures of various traffic events. To this end, we introduce a novel dataset, SUTD-TrafficQA (Traffic Question Answering), to facilitate the research of causal reasoning in complex traffic scenarios.

In our dataset, to help develop models for addressing several major and concerning issues in intelligent transportation, we design 6 challenging reasoning tasks, which require exploring the complex causal structures within the inference process of the traffic events. 
As shown in Figure \ref{fig:example},
these tasks correspond to various traffic scenarios involving both road-agents and surroundings, and the models are required to forecast future events, infer past situations, explain accident causes, provide preventive advice, and so on. 

To present these reasoning tasks, video question answering \cite{zhu2017uncovering} is a natural and effective choice, and is used for our dataset construction, since to accurately answer the given questions, the models need to acquire strong capabilities of performing various levels of logical reasoning and spatio-temporal cognition for the events.

Besides providing the challenging and useful reasoning tasks, 
we adopt a combination scheme of online collection and offline capturing to collect videos, such that the data in our benchmark covers various traffic events, diversified road-agents and surroundings, and different capturing perspectives in the wild. With the provided various tasks and diverse videos, our dataset shall be able to serve as a comprehensive benchmark for video reasoning of traffic events.

In some application scenarios, (e.g., assisted driving
), the computational resource and energy budget can be constrained.
Thus 
both the inference accuracy and the computation efficiency are important for video event reasoning in these scenarios.
Existing video QA methods \cite{Kim_2020_CVPR,Le_2020_CVPR,lei-etal-2020-tvqa} mainly focus on strengthening the reasoning accuracy without emphasizing much efficiency, and most of
these works apply fixed computation pipelines 
to answer different questions, while ignoring to conduct adaptive and efficient computation resource allocation based on the logic structure behind reasoning over video events. 

%

In this paper, to achieve reliable and efficient video reasoning, we propose \textbf{Eclipse}, an \textbf{E}ffi\textbf{c}ient g\textbf{li}m\textbf{pse} network.
Specifically, considering there is often large redundancy among video frames, via dynamic inference, our network adaptively determines where to skip and glimpse at each step, and what computation granularity needs to be allocated for the glimpsed frame. Such a dynamic reasoning scheme avoids feature extraction for the irrelevant segments in the video, and hence significantly reduces the overall computation cost towards 
reliable and efficient reasoning.
%
It is noteworthy that both the determination of selecting a glimpse frame and the decision of computation granularity for each glimpse are essentially discrete operations, which are 
not trivial to optimize. To handle this issue, an effective joint Gumbel-Softmax mechanism is also introduced in this paper, which makes our Eclipse framework fully differentiable and end-to-end trainable. 

To the best of our knowledge, this is the first work that simultaneously performs adaptive frame localization and feature granularity determination in a novel dynamic reasoning process for reliable and efficient causal reasoning and video QA. A joint Gumbel-Softmax operation is also introduced in this work to optimize the two decisions jointly.


\section{Related Works}
\textbf{Intelligent Transportation.} 
With the rapid development of deep learning techniques \cite{krizhevsky2012imagenet, ren2015faster, vaswani2017attention}, data-driven intelligent transportation \cite{chandra2020forecasting,lou2019veri,you2020traffic,chandra2020forecasting} has emerged as a prominent research topic.
Lou et al. \cite{lou2019veri} presented a dataset together with an adversarial learning model for vehicle re-identification.
%
Different from existing intelligent transportation datasets and methods,
in this paper, we investigate the problem of causal reasoning with video QA over various traffic scenarios. A new benchmark, SUTD-TrafficQA, together with a novel model, Eclipse, is proposed for this challenging task.

\begin{table*}[t]
\caption{Comparison among SUTD-TrafficQA and some other video QA datasets. Providing challenging \textbf{traffic-scenario reasoning tasks} with \textbf{real-world videos} and \textbf{human-generated QA pairs}, our dataset shall serve as a comprehensive and challenging benchmark for video reasoning over traffic events.}
\vspace{-0.2cm}
\begin{center}
\scriptsize
\setlength\tabcolsep{0.5pt}
\begin{tabular}{c|c|ccccccc} \hline
\multirow{2}{*}{\textbf{Dataset}} & \textbf{Synthetic Videos} & \multicolumn{7}{c}{\textbf{Real-World Videos}} \\
\cline{2-9}
& CLEVRER \cite{yi2019clevrer} 
        & MovieQA \cite{tapaswi2016movieqa} & MSRVTT-QA \cite{xu2017video} & TGIF-QA \cite{jang2017tgif} & TVQA \cite{lei2018tvqa} &  MarioQA \cite{mun2017marioQA} & Social-IQ \cite{zadeh2019social} &  \textbf{SUTD-TrafficQA (Ours)} \\ \hline
Basic Understanding  & $\checkmark$ & $\checkmark$ & $\checkmark$ & $\checkmark$ & $\checkmark$ & $\checkmark$ & $\checkmark$ & \textbf{$\checkmark$} \\ 
Attribution  & $\checkmark$ & $\checkmark$ & $\times$ & $\times$ & $\checkmark$ & $\checkmark$ & $\checkmark$ & \textbf{$\checkmark$} \\
Event Forecasting & $\checkmark$ & $\times$    & $\times$ & $\times$     & $\times$     & $\times$     & $\times$     & \textbf{$\checkmark$}  \\ 
Reverse Reasoning  & $\times$     & $\times$     & $\times$ & $\times$     & $\times$     & $\times$     & $\times$     & \textbf{$\checkmark$} \\ 
Counterfactual Inference & $\checkmark$ & $\times$    & $\times$ & $\times$     & $\times$     & $\times$     & $\times$     & \textbf{$\checkmark$} \\ 
Introspection   & $\times$     &$\times$     & $\times$&  $\times$     & $\times$     & $\times$     & $\times$     & \textbf{$\checkmark$} \\
\hline
QA Generation & Automatic & Human & Automatic & Automatic \& Human & Human & Automatic & Human & Human \\ \hline
Topic & Synthetic Object Collision & Movies & Various Scene & GIFs & TV-Shows & Gameplay & Social Behavior & Traffic Events\\ \hline
\end{tabular}
\end{center}
\label{table:dataset_comparison}
\vspace{-0.5cm}
\end{table*}

\textbf{Video QA Datasets.}
Recently, there emerges a great interest in visual reasoning and question answering in videos \cite{10.1007/s11263-016-0987-1,maharaj2017dataset,zhu2017uncovering,ijcai2017-280}, and several video QA datasets \cite{xu2017video,zeng2016leveraging,tapaswi2016movieqa,zadeh2019social,jang2017tgif,garcia2020knowit,emrvqasongMM18,yu2019activityqa} have been developed. Among them, 
%
MovieQA \cite{tapaswi2016movieqa} and TVQA \cite{lei2018tvqa} present the movie and TV-show videos respectively with human-generated questions. More recently, 
CLEVRER \cite{yi2019clevrer} focuses on collision event reasoning among several simple visual objects
in a controlled environment using fully synthetic videos. 
Differently, our SUTD-TrafficQA focuses on reasoning over the complex traffic scenarios in the wild, 
where 6 challenging tasks for traffic event reasoning are introduced based on the diverse real-world
traffic
videos.  

\textbf{Video QA Methods.}
Extensive studies have been conducted for video QA \cite{yu2017end,jang2017tgif,garcia2020knowledge,Jiang_2020_CVPR,Kim_2020_CVPR,lei-etal-2020-tvqa,tsai2019GSTEG,li2019beyond,Kim_2018_ECCV,Yu_2018_ECCV,sukhbaatar2015end,wang2018movie,fan2019heterogeneous,liang2018focal,tensoremnlp17,8654010}. 
Yu et al. \cite{yu2017end} employed LSTM to encode videos and QA pairs, and adopted an attention mechanism \cite{you2016image}. 
Jang et al. \cite{jang2017tgif} used LSTMs with a different attention scheme to capture spatio-temporal patterns in videos. 
Different from existing video QA methods, our Eclipse model investigates the direction of learning an effective glimpse policy for adaptive reasoning to achieve reliable reasoning with computation efficiency.

\textbf{Computation-Efficient Models.}
Recent works \cite{Bhardwaj_2019_CVPR,Mu_2019_CVPR,strubell2019energy,figurnov2017spatially, korbar2019scsampler,wu2019adaframe,schwartz2019green,NEURIPS2019_bd853b47, fan2020adaptive} have pointed out the need of improving the computation-efficiency when designing deep models, and 
different strategies, 
including filter pruning \cite{li2016pruning}, weight sparsification \cite{sun2016sparsifying}, vector quantization \cite{agustsson2017soft}, and dynamic routing \cite{wang2018skipnet}, etc., have been proposed. 
In this paper, we 
propose an efficient model, Eclipse, the first model that performs dynamic inference and adaptive computation adjustment for video QA, which leverages an effective glimpse policy with the guidance of text and visual context information for both glimpse frame selection and computation granularity determination.

\section{SUTD-TrafficQA Dataset}\label{section:dataset}

Our dataset contains 62,535 QA pairs and 10,080 videos of traffic scenes. Below we first propose 6 challenging traffic-related reasoning tasks, and then introduce the QA collection process and the dataset statistics. 

{\bf Basic understanding.} This task evaluates the ability of the models in perceiving and understanding traffic scenarios at the basic level, which consists of multiple sub-tasks including feature-query (e.g., vehicle type, road situation, and environment description), event-query (e.g., accident existence, pedestrian action analysis, and events temporal relation), event classification (e.g., accident type), and counting (e.g., road-agent number). 

{\bf Event forecasting.} This task requires a model to infer future events based on observed videos, and the forecasting questions query about the outcome of the current situation. 

{\bf Reverse reasoning.} This task is to ask about the events that have happened before the start of a video segment.
%

{\bf Counterfactual inference.} This task queries the consequent outcomes of certain hypothesis (e.g., what if the blue sedan had not accelerated?). The hypothetical conditions do not occur in the video, so the model needs to reason about the imagined events under the designated condition.

{\bf Introspection.} This task is to test if models are able to provide preventive advice (e.g., what could the pedestrian have done to avoid the collision with the car?). 
The candidate answers list actions that could have been taken to avoid traffic accidents or congestion.

{\bf Attribution.} This task seeks the explanation about the causes of traffic events (e.g., what are the reasons of the rear-end crash?), so as to check if models are able to infer the underlying factors leading to the event. 

We define all the above reasoning tasks as multiple-choice questions without limiting the number of candidate answers for each question. The number of candidate answers varies from 2 to 12 for different questions. 
We then sample among the candidate answers to balance the dataset 
and limit the occurrence of the same correct answers within each task to minimize language biases. As summarized in Table \ref{table:dataset_comparison}, by introducing these challenging tasks, our dataset complements existing datasets, and facilitates the exploration of video QA in complex traffic scenarios.

\subsection{QA Collection}

{\bf Videos.} We collected videos by using a combination of online harvesting and offline capturing, to cover various real-world traffic scenarios.
As for online video collection, a variety of video sharing platforms, based in different countries, are used to increase the diversity, including but not limited to YouTube, LiveLeak, Twitter, and Bilibili. More than nine thousand videos were thus collected from these online sources.
As for offline video capturing, a set of videos were captured via handheld cameras by volunteers, while another set were fetched from car-mounted video recorders. These two sets of offline videos were then examined and trimmed into around one thousand video clips.

After combing the online videos and offline captured ones, we obtain a total of $10,080$ videos containing diversities in various aspects, including: a) different weathers (sunny/rainy/windy/snowy); b) different time (daytime/night); c) diverse road situations (congested/sparse, urban/rural roads); d) various traffic events (accidents, vehicle turning, pedestrian behaviors, traffic lights, etc.); e) different video perspectives (surveillance camera perspective/car-mounted video perspective/hand-held camera perspective); and f) various clip lengths (from $1$ to $70$ seconds).

{\bf QA pairs.} As the first step, the 6 tasks were explained to annotators. To ensure they fully understand the design principle of the tasks, multiple example questions were prepared for them to identify which task the questions belong to. Afterwards, each annotator was presented with batches of video folders, where each folder contains 100 clips that were randomly selected from the full video set. Annotators were asked to create at least 3 questions of the reasoning tasks for each video.
We did not impose constraints on question formats to encourage annotators to keep their QA pairs diversified. Specifically, we encourage them to rephrase similar questions or candidate answers to push the models to learn underlying semantics of QA pairs rather than superficial language correlation. 
To ensure the quality of QA pairs, 
we cross-checked QA annotations on a weekly basis. 
In addition, we kept monitoring the distribution of different tasks in QA collection to maintain the balance and diversity of our dataset.

\subsection{Dataset Statistics}

\begin{figure}[h]
\begin{center}
    \includegraphics[width=1\linewidth]{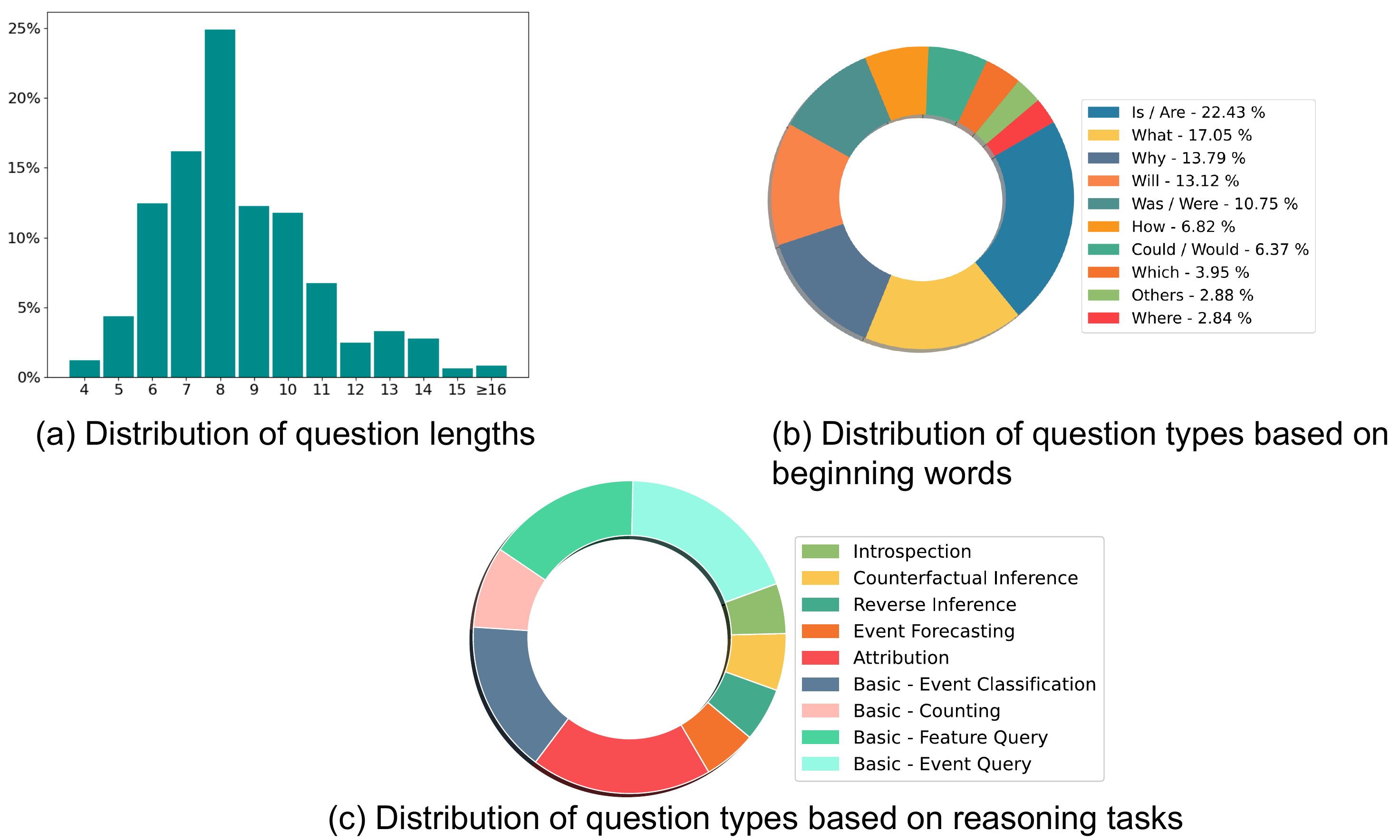}
\end{center}
\vspace{-0.4cm}
\caption{
Statistics of SUTD-TrafficQA. More in supplementary.
}
\label{fig:Dataset_stats}
\vspace{-0.2cm}
\end{figure}

In this part, we present the statistics of our SUTD-TrafficQA dataset. Figure \ref{fig:Dataset_stats} (a) demonstrates the distribution of question length measured by number of words. The average length of questions is $8.6$ words. Figure \ref{fig:Dataset_stats} (b) shows the various question types categorized by their beginning words, which implies the diversity of questions in our dataset.
Figure \ref{fig:Dataset_stats} (c) presents the split of questions in terms of reasoning tasks.  
Our dataset covers a broad range of traffic-related reasoning tasks requiring various levels of spatio-temporal understanding and causal reasoning in videos. 
\section{Eclipse Network}

To deal with video reasoning, a common solution is to watch the full video and analyze the whole event information carefully. In this manner, generally, a fixed computation architecture \cite{lei2018tvqa,jang2017tgif,arad2018compositional,Le_2020_CVPR} can be applied over the whole video for tackling each question. 
However, using a fixed network architecture for handling the video QA task is often computation-heavy and energy-consuming \cite{strubell2019energy,schwartz2019green}, because the video sequence used for reasoning can be very long and contain plenty of frames for processing.


Recalling that, as humans,
to analyze events in a video, we may not be patient enough to scrutinize the frames in the whole video,
instead, we may 
adopt an adaptive information ``foraging'' strategy  \cite{fitzsimmons2014skim,duggan2011skim}.
Concretely, we may 
use a ``dynamic'' inference manner 
to skip forth and back over the sequence to progressively infer and select some useful frames based on the task. 
Moreover, for the picked frames, we may examine a few of them very carefully while glancing over others. 
Such a dynamic and adaptive perception habit \cite{fitzsimmons2014skim,duggan2011skim} 
frees us from watching the whole video thoroughly, and often enables fast yet still very accurate video reasoning at a very small frame usage.

\begin{figure}[tbp]
\begin{center}
    \includegraphics[width=1\linewidth]{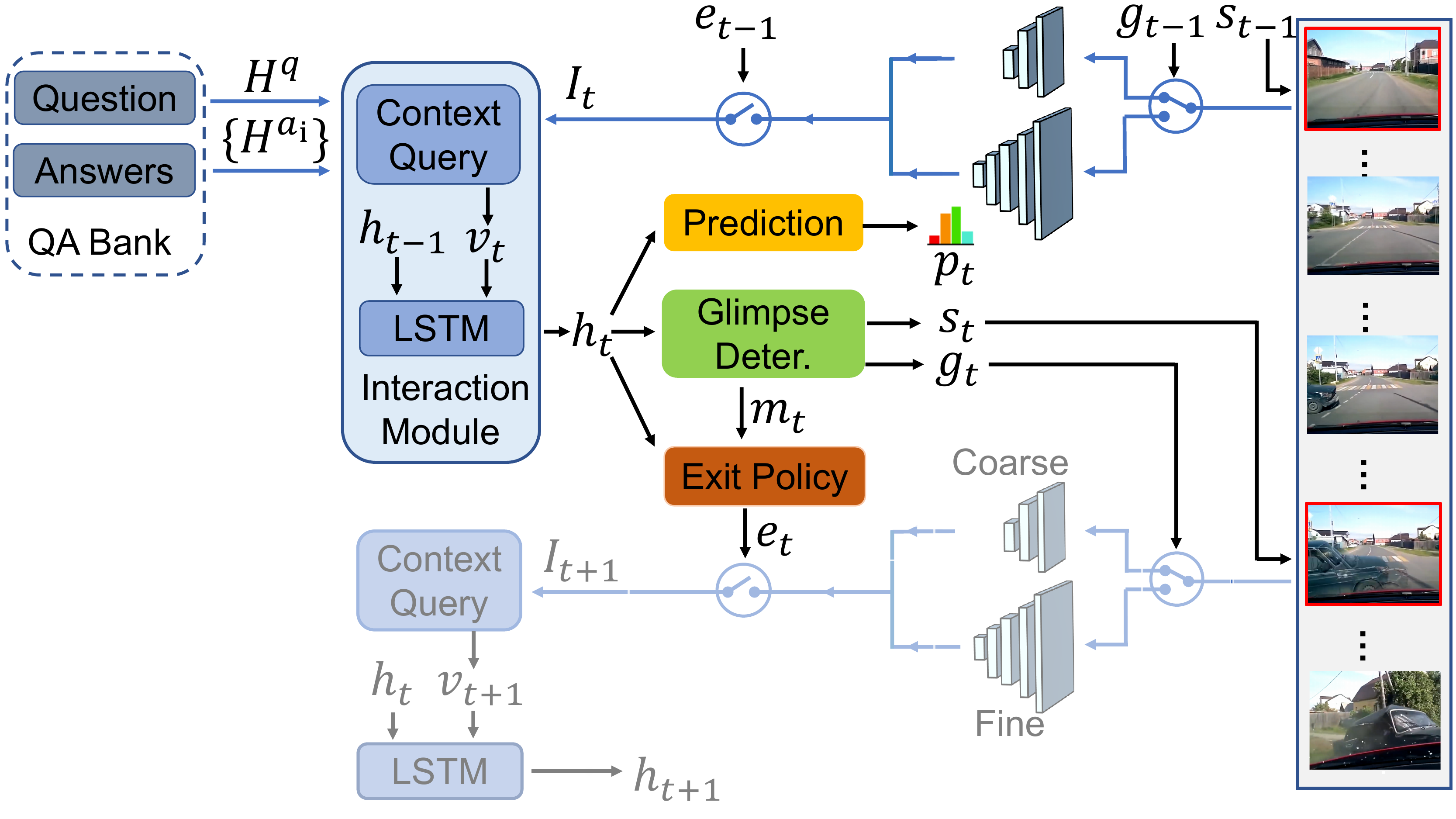}
\end{center}
\vspace{-0.5cm}
\setlength\abovecaptionskip{-2cm}   
\setlength\belowcaptionskip{-2cm}
\caption{Architecture of Eclipse for dynamic causal reasoning.}
\label{fig:Eclipse}
\vspace{-0.4cm}
\end{figure}

Motivated by this, we aim to explore the direction of efficient and dynamic reasoning in complex traffic scenarios. Thus, we propose an \textbf{E}ffi\textbf{c}ient g\textbf{li}m\textbf{pse} (\textbf{Eclipse}) network for video QA, as illustrated in Figure \ref{fig:Eclipse}.
Instead of using a fixed computation architecture over each video and question, our network learns to dynamically skip to and select a useful video frame at each inference step. Moreover, our network adaptively decides the feature computation granularity (i.e., coarse or fine) of the selected frame. To perform such a process, at each inference step, our network takes advantage of the guidance information including the QA pair, the currently selected frame and the historical cues for dynamic reasoning.

Specifically, as shown in Figure \ref{fig:Eclipse},
in our network, to provide QA information for the Interaction Module, the QA Bank stores the representation of the QA pairs. 
At each inference step, to assist the dynamic reasoning of selecting the frame and the corresponding feature granularity, the Interaction Module leverages the QA information, the currently selected frame and the information from historically observed frames to derive an expressive representation, 
which then serves as the input of the dynamic reasoning process performed by the downstream modules, including the Prediction Module for outputting the reasoning result, and
Glimpse-Determination Module for dynamically determining which frame to be observed at next step. 
Besides, the Exit-Policy Module also uses this representation to adaptively decide whether we can exit the reasoning process at current inference step. 
%
Via such a dynamic and recurring reasoning process, our network can derive reliable answer predictions with notable computation efficiency w.r.t. both the frame usage and feature computation granularity. We elaborate the network modules in detail below.

{\bf QA Bank.} To provide the QA information as the guidance for dynamic reasoning, our QA Bank encodes the representation of the question via a bi-directional LSTM. 
By concatenating hidden states of the BiLSTM, the question representation can be denoted as $H^q\in\mathbb{R}^{n_q\times2d}$, where $n_q$ is the number of words in the question, $d$ denotes the dimension of LSTM hidden state. Similarly, we encode all the candidate answers as
${\{ H^{a_i} \}}^N_{i=1}$,
where $H^{a_i}\in\mathbb{R}^{n_{a_i}\times2d}$,
$N$ is the number of candidate answers, 
and $n_{a_i}$ is the number of words in the answer $a_i$.

{\bf Interaction Module.} To assist the subsequent dynamic reasoning with rich guidance information, we design an Interaction Module to fuse different kinds of available inputs. This module, consisting of a context-query sub-module and an interaction LSTM, recurrently interacts with a small number of frames selected from the full video sequence containing $T$ frames. As shown in Figure \ref{fig:Eclipse}, at each inference step $t$, this module fuses the QA information ($H^q$ and $\{H^{a_i}\}$),
the currently selected frame ($I_t$), and the historical cues ($h_{t-1}$)
to produce an expressive representation ($h_t$) that can be used for dynamic reasoning. 

More formally, at the $t^{th}$ inference step, to first fuse the textual QA information ($H^q$ and $\{H^{a_i}\}$) with the currently selected visual frame feature ($I_t$),
we use a context-query sub-module to perform such a fusion process. We implement this sub-module by following the context-matching module of \cite{lei2018tvqa}, which can effectively fuse the visual feature sequence with the textual sequence to produce a combined representation. Since the original method in \cite{lei2018tvqa} takes a sequence of visual features as an input, and we have only the selected frame feature $I_t$ here, we treat $I_t$ as a visual feature sequence containing a single element. Hence the fusion process in this sub-module can be formulated as:
\begin{equation}\label{eq.context-query}
    v^i_t=[I_t;F^{I_t,q};F^{I_t,a_i};I_t\odot F^{I_t,q};I_t\odot F^{I_t,a_i}]
\end{equation}
where $\odot$ represents element-wise product. $F^{I_t,q}$ and $F^{I_t,a_i}$ are obtained by computing the similarity between the visual frame feature ($I_t$) and the textual features ($H^q$ and $H^{a_i}$). More details of this process are referred to \cite{lei2018tvqa} and also our supplementary.
For simplicity, we use $v_t$ to represent the concatenated ${\{v^i_t \}}^N_{i=1}$, as shown in Figure \ref{fig:Eclipse}. The 
output $v_t$, that incorporates the information of currently selected frame and the QA embedding, can be fed into the following interaction LSTM for more information interaction, as introduced below.

Besides fusing the QA information and the currently selected frame ($I_t$), to guide the dynamic reasoning at current inference step, we also incorporate the historical cues from 
past inference steps. Thus we design an interaction LSTM that takes the $v_t$ as the new input to interact with the historical cues $h_{t-1}$ as:
\begin{equation}\label{eq.lstm}
    c_t, h_t = LSTM(v_t, c_{t-1}, h_{t-1};\theta_{LSTM})
\end{equation}
where $\theta_{LSTM}$ are parameters of LSTM. The generated hidden state $h_t$ encodes rich information of all available inputs and historical cues, and thus it can serve as an expressive representation to be fed into the downstream modules for dynamic reasoning as follows.

{\bf Prediction Module.} This module is used to generate the reasoning result (i.e. the probability distribution over candidate answers) at the current inference step $t$. This module computes the reasoning result as: $p_t=f_{p}(h_{t};\theta_{p})$, where $p_t\in\mathbb{R}^{N}$ represents the probability scores for all candidate answers. $f_{p}$ can be implemented with one fully-connected (FC) layer followed by a $Softmax$ classifier.

{\bf Glimpse-Determination Module.} At each inference step $t$, conditioned on $h_t$, this module performs dynamic reasoning by making two decisions simultaneously. The first decision is to select which frame to be observed at next step, and the second is to decide whether to compute fine-grained features or coarse features for this selected frame. 
Corresponding to these two decisions, we design the following two branches within this module.

The skip-policy branch selects the frame that we need to skip to at next inference step via the following process: $s_t=f_{s}(h_{t};\theta_{s})$, where the output $s_t$ indicates the decision of the next frame location. Note that our network can skip forth and back over the entire video sequence, which is conceptually similar to the human reasoning process where we need to not only jump forward to find future informative frames but also go back to examine past information. 

Besides determining next frame, this module also has a granularity-policy branch that adaptively decides the feature computation granularity for the next selected frame, formulated as:
$g_t=f_{g}(h_t;\theta_{g})$. The output $g_t$, 
denotes the decision of feature granularity. In our implementation, we provide two kinds of feature granularity, namely, coarse features computed by a lightweight CNN; fine-grained features computed by a more representative yet computation-heavier CNN, to be chosen from. In the Glimpse-Determination module, both $f_{s}$ and $f_{g}$ are implemented with a FC layer. 


{\bf Exit-Policy Module.} To estimate when we can exit the reasoning process to achieve adaptive inference, we design the Exit-Policy Module. At each inference step $t$, this module decides if we can exit the reasoning process at current step based on the guidance information ($h_t$) as: $e_{t}=f_{e}(h_{t};\theta_e)$, where the output $e_t$ denotes
the confidence score of terminating the reasoning at current step. By training the exit-policy, 
our network can achieve adaptive inference, such that only a small and flexible number of frames are selected and computed on a per-video basis to derive reliable reasoning result.

{\bf Optimization.} To optimize the above modules in our network, we introduce several loss functions. Specifically, at each inference step $t$, a cross-entropy loss $\mathcal{L}^t_{pred}$ is used to train the classifier of the Prediction Module:
\begin{equation}\label{eq.cross-entropy loss}
    \mathcal{L}^t_{pred}= -\sum_{n=1}^{N} y^n \log (p^{n}_{t})
\end{equation}
where $y$ is the ground-truth one-hot label vector for the candidate answers and $p^{n}_{t}$ is the predicted score for the $n^{th}$ answer.
As for Glimpse-Determination Module, to push the skip-policy branch to select a useful frame at each inference step, a simple yet effective loss is used:
\begin{equation}\label{eq.increment_loss}
    \mathcal{L}^t_{incre} = -(m_{t} - m_{t-1}) 
\end{equation}
where $m_{t}=p_{t}^{gt}-\max \{ p_{t}^{c'}\mid c'\neq gt \}$ is the margin between the predicted probability of the correct answer (indexed by $gt$) and the largest probability of other candidate answers. We can simply infer that a larger $m_t$ indicates a more confident and accurate reasoning. Therefore, we use $m_{t}-m_{t-1}$ to encourage the margin to keep growing over the inference steps, which indicates that at each step, we aim to select a useful frame to benefit our dynamic reasoning, considering that the confidence of our network for the correct answer increases when seeing the selected frames.

Meanwhile, to further save computation cost and prevent the granularity-policy branch from constantly using the computation-heavy fine features, we penalize the feature granularity policy
when the fine feature is computed at each inference step $t$ as follows:
\begin{equation}\label{eq.feature_loss}
    \mathcal{L}^t_{feat}= g_{t}
\end{equation}
where $g_t=1$ represents that the policy chooses to extract computation-heavy fine features for next step, while $g_t=0$ means it switches to extract computation-cheap coarse features for next step.
To optimize our whole Glimpse-Determination Module, we incorporate the above two loss functions,   $\mathcal{L}^t_{incre}$ and $\mathcal{L}^t_{feat}$, into a combined loss:
\begin{equation}\label{eq.glimpse_loss}
    \mathcal{L}^t_{glimpse}= \mathcal{L}^t_{incre} + \mathcal{L}^t_{feat}
\end{equation}

Last but not least, we need to train the Exit-Policy to make a reliable decision if we can exit the reasoning process at current inference step. However, there are no ground-truth labels providing feedback on when our network can exit reasoning.
Therefore, we leverage $m_t$ 
to generate dynamic labels to train Exit-Policy. Recalling that $m_{t}$ is the probability margin between the correct answer and the largest one of other candidate answers at each step, and $m_{t}$ is optimized to keep increasing 
under the constraint of Eqn. \ref{eq.increment_loss}. 
Therefore, the gap between $m_t$ at different inference steps can be used to estimate the information gain of seeing more frames in our dynamic reasoning process. Given the pre-defined largest reasoning step $T$, the gap between $m_t$ (at the current step) and $m_T$ (of the final step) can estimate the value of remaining information gain by continuing reasoning till the end. Thus the gap can be used to determine whether our network can stop inference at $t^{th}$ step. When $m_t$ is very close to $m_T$, this means the information gain by observing more frames is small, and thus we can exit reasoning in advance to reduce computation without incurring decrease in prediction accuracy.

In particular, at each step $t$, if $m_{T} - m_{t}<\mu(m_{T}-m_1)$, which means $m_t$ is close to $m_T$, i.e., the estimated remaining information gain is small enough, we set the label, $y^t_{exit}$, as 1, indicating our model can exit reasoning at the $t^{th}$ inference step. Otherwise, the label is set to 0, representing we need to seek more frames. Here $\mu>0$ controls how close $m_t$ should be to $m_T$ when the network exits inference. Conditioned on the estimated labels, $y^t_{exit}$, training Exit-Policy can be seen as a binary classification problem. Thus we train this module by minimizing a binary cross-entropy loss:
\begin{equation}\label{eq.bce_loss}
    \mathcal{L}^t_{exit}= -[y^t_{exit}\log(e_{t}) + (1-y^t_{exit})\log(1-e_{t})]
\end{equation}
By combining Eqns (\ref{eq.cross-entropy loss}), (\ref{eq.glimpse_loss}),
and (\ref{eq.bce_loss}), the total loss function for each step $t$ can be formulated as:
\begin{equation}\label{eq.final_loss}
    \mathcal{L}^t = \mathcal{L}^t_{pred} + \mathcal{L}^t_{exit} + \lambda *\mathcal{L}^t_{glimpse}
\end{equation}
where $\lambda$ is the weight of the combined loss function for optimizing the Glimpse-Determination Module.
In our experiments, we compute the sum: $\sum_{t=1}^{T}{\mathcal{L}^t}$ from all inference steps as the final optimization objective. 

Note that Eqn. (\ref{eq.final_loss}) cannot be optimized directly with gradient descent, since the involved decisions of selecting frames and determining feature granularity in our dynamic reasoning are discrete, and sampling from discrete distribution makes the network non-differentiable. To address this issue, we introduce an effective joint Gumbel-Softmax operation.

{\bf Joint Gumbel Softmax.} The original Gumbel-Softmax Sampling \cite{jang2016categorical} is an effective way to transform the original non-differentiable sample from a discrete distribution, to a differentiable decision from a corresponding Gumbel-Softmax distribution. In our task, to sample from the aforementioned two discrete distributions (namely, selecting frames and determining granularity) simultaneously, we here design an effective joint Gumbel-Softmax operation. 

In particular, in the Glimpse-Determination Module, at each step $t$, we first derive the logits $z \in\mathbb{R}^{T*2}$ by feeding the hidden state $h_t$ into a fully-connected layer. 
Then we use $Softmax$ to obtain a categorical distribution $\pi_t$ from $z$:
    $\pi_t = \left\{ p_{i,j} \mid p_{i,j}=\frac{\exp(z_{i,j})}{\sum_{c=1}^T\sum_{k=1}^2\exp(z_{c,k})} \right\}$.
With the Gumbel-Max trick \cite{jang2016categorical}, the discrete sample from the categorical distribution $\pi_t$ can be defined as follows:
\begin{equation}\label{eq.argmax}
    \hat{l_t} = \mathop{\arg\max}_{i\in\{1,...,T\},j\in\{1,2\}} (\log p_{i,j} + g_{i,j})
\end{equation}
where $g_{i,j} = -\log (-\log(u_{i,j}))$ denotes the Gumbel noise, and $u_{i,j}$ is the i.i.d. samples drawn from $Uniform(0,1)$. We can further relax the non-differentiable operation $argmax$ with $softmax$ to facilitate gradient-based optimization:
\begin{equation}\label{eq.argmax_frame_relaxation}
\begin{aligned}
  l_t = &\left\{ P_{i,j} \mid P_{i,j}=\frac{\exp{((\log p_{i,j} + g_{i,j})/\tau)}}{\sum_{c=1}^{T}\sum_{k=1}^{2}\exp{((\log p_{c,k} + g_{c,k})/\tau)}}\right\},\\ 
    &for\ i\in\{1,...,T\}, j\in\{1,2\} 
\end{aligned}
\end{equation}
where $\tau$ is the temperature parameter, which controls the smoothness of the sampling mechanism. When $\tau\to0$, the sampling approximates the $argmax$ operation in Eqn. \ref{eq.argmax}. The output $l_t$ incorporates the output of two decisions: the first dimension of $l_t$ denotes the decision of selecting the frame (i.e, the output $s_t$ of the skip-policy branch) and the second dimension of $l_t$ denotes the decision of the feature granularity of the selected frame (i.e, the output $g_t$ of the granularity-policy branch).

By using the outputs ($s_t$ and $g_t$) of the joint Gumbel-Softmax operation, our network manages to dynamically select the frame for next inference step and specify the feature granularity for the selected frame at each step. 
Therefore, by introducing the joint Gumbel-Softmax, our network can learn the two discrete policy decisions jointly in a fully differentiable way.

{\bf Training and testing.} During training, we optimize our network within a fixed number of steps, which means the exit-policy is trained together with other modules but the exit decisions 
are not used. However, at the testing phase, if the exit-policy decides to stop reasoning at the $t^{th}$ inference step, we exit the model and use the current prediction result, $p_t$, as the final reasoning result. In such a manner, our model achieves dynamic causal reasoning.

\section{Experiments} \label{sec:exp}
Given that the number of candidate answers for each question is not fixed in our dataset, we evaluate the performance of our network using binary and multi-choice setups. In binary case (denoted as \textbf{Setting-1/2}), the input to the model is a question with an answer, and the model needs to predict the correctness of this answer. In multi-choice setup (denoted as \textbf{Setting-1/4}), models are expected to select the correct answer from 4 candidate answers (i.e, 3 of them are incorrect). These two experiment setups can be treated as binary and four-class classification problems.

{\bf Implementation Details.} 
We compute features from the penultimate layer of a pretrained ResNet-101 model \cite{he2016deep} as the fine-grained frame feature,
and a pretrained MobileNetv2 \cite{sandler2018mobilenetv2} is used as the lightweight CNN to extract coarse features. In the QA Bank, we use Glove \cite{Pennington14glove:global} to embed QA text, and then use a BiLSTM with 150-Dimension hidden states to encode the textual sequence.  As for the Interaction LSTM, the dimension of hidden states is 300.
We implement the framework using Pytorch and adopts Adam \cite{kingma2014adam} with a learning rate of 3e-4 and a weight-decay of 1e-5. The $\mu$ in the Exit-Policy is set to 0.1 and $\lambda$ is set to 0.01 in the loss function. We follow \cite{jang2016categorical} and set the initial temperature $\tau$ to 5, and gradually anneal it with an exponential decay factor of -0.045 in every epoch. According to evaluation statistics, our network shows a very fast \textbf{inference speed} of $16ms$ per testing video on a Nvidia RTX 2080Ti GPU.

We compare our network with the following baselines.
{\bf Text-only models.} These models only relying on text information without visual input, are relatively weak baselines used to assess language biases in our SUTD-TrafficQA. {\bf Q-type (random)} randomly selects an answer from the answer space. {\bf QE-LSTM} uses Glove \cite{Pennington14glove:global} to embed the input question and then encode it with LSTM \cite{hochreiter1997long}. The final LSTM hidden state is passed to a MLP for predicting the correct answer. Different from {\bf QE-LSTM} using questions only, {\bf QA-LSTM} uses LSTM to encode both question embedding and answer embedding, and the final hidden states are used for predicting the answer.

{\bf Text+video models.} We evaluate the following models that require both video and text inputs. 
{\bf VIS+LSTM} \cite{ren2015exploring} uses LSTM to encode image representation and textual features. Since the original method takes a single image as input, we adapt this method by averaging features of all sampled frames in a video as the visual input. 
{\bf Avgpooling} uses each frame with the encoded QA features to compute a prediction for each frame.
We then perform mean pooling over all the frame predictions to obtain the final result. 
{\bf CNN+LSTM} uses two LSTMs to encode both the video sequence and the QA text respectively. The two final hidden states are concatenated to predict the correct answer.
{\bf I3D+LSTM} uses I3D network \cite{carreira2017quo} to extract video motion features, and then fuse with QA textual features encoded by LSTM to compute the model prediction.
{\bf TVQA} \cite{lei2018tvqa} is a multi-stream network to fuse input features from different modalities to answer the question. 
{\bf HCRN} \cite{Le_2020_CVPR} adopts a Hierarchical Conditional Relation Networks to model sophisticated structure for reasoning over videos.
{\bf BERT-VQA} \cite{Yang_2020_WACV} uses BERT \cite{devlin2018bert} 
to encode the visual and language information jointly to predict the answer. 
{\bf }


\subsection{Results and Analysis}

\begin{figure*}[t]
\setlength\abovecaptionskip{-5cm}
\setlength\belowcaptionskip{-5cm}
\begin{center}
    \includegraphics[width=1\linewidth]{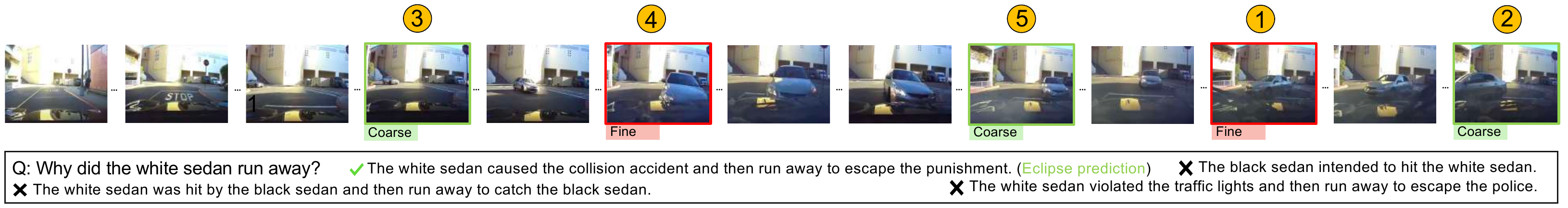}
\end{center}
\vspace{-0.3cm}
    \caption{A qualitative example. The numbers above the selected frames show the order of the sequence selected by our network. It shows that our model selects informative frames dynamically and allocates large computation budget using fine features to most relevant frames for causal reasoning. More examples in supplementary.}
\label{fig:policy_example}
\vspace{-0.3cm}
\end{figure*}

\begin{table}[t]
\vspace{-0.2cm}
\caption{Results on SUTD-TrafficQA dataset. 
}
\setlength\abovecaptionskip{-2cm}
\setlength\belowcaptionskip{-2cm}
\setlength{\tabcolsep}{10pt}
\scriptsize
\begin{center}
\begin{tabular}{l|ccc}
\hline
Models & Setting-1/4 & Setting-1/2 & GFLOPs  \\\hline
Q-type (random)& 25.00 & 50.00 & - \\ 
QE-LSTM & 25.21 & 50.45 & - \\ 
QA-LSTM & 26.65 & 51.02 & -  \\ \hline
Avgpooling & 30.45& 57.50 & 252.69 \\ 
CNN+LSTM & 30.78 & 57.64 & 252.95  \\ 
I3D+LSTM & 33.21 & 54.67 & 108.72  \\
VIS+LSTM \cite{ren2015exploring} & 29.91 & 54.25 & 252.80\\
BERT-VQA \cite{Yang_2020_WACV} & 33.68 & 63.50 & 266.77 \\ 
TVQA \cite{lei2018tvqa} & 35.16 & 63.15 & 252.11  \\ 
HCRN \cite{Le_2020_CVPR} & 36.49 & 63.79 & 2051.04 \\ 
\textbf{Eclipse}  & \textbf{37.05} & \textbf{64.77} & \textbf{28.14} \\ \hline
\textit{Human}    &  95.43    & 96.78 & -\\ \hline
\end{tabular}
\end{center}
\label{table:baseline_comparison}
\vspace{-0.5cm}
\end{table}

The results in Table \ref{table:baseline_comparison}
demonstrate the minimal language biases in SUTD-TrafficQA, as the text-only baselines perform almost the same as the random choice. In contrast, the models using video input achieve obviously higher accuracy than text-only baselines. This demonstrates that to solve the reasoning tasks in our dataset, the model needs to associate visual content with linguistic cues to infer correct answers. For a fair computation-efficiency comparison, we pick models requiring video input to compute GFLOPs per video, as the visual feature extraction consumes much computation budget, and the metric of GFLOPs is independent of hardware configurations. The results show our Eclipse achieves state\text{-}of\text{-}the\text{-}art reasoning accuracy 
with significantly improved computation efficiency. This verifies that compared to conventional video QA  methods, through dynamic causal reasoning, our model effectively exploits the spatio-temporal and logical structure of video events to infer correct answers with much smaller frame usage and efficient feature computation.
In addition, three volunteers who did not see the videos and questions before, were invited to pick correct answers, and we use the average prediction accuracy as \textit{Human} performance. The discrepancy between neural networks and the human performance demonstrates the challenging nature of our dataset and the necessity of further research in video reasoning area.

\begin{table}[tbp]
\caption{Results of removing granularity-policy in Eclipse.
}
\vspace{-0.2cm}
 \setlength{\tabcolsep}{0.5pt}
\scriptsize
\begin{center}
\begin{tabular}{c|c|c|c}
\hline
Models & Coarse Features Only & Fine Features Only & ~Eclipse~(dynamic granularity) \\\hline
Accuracy & 34.35 & 37.16 & 37.05 \\ \hline
GFLOPs & 10.61 & 133.75 & 28.14 \\ \hline
\end{tabular}
\end{center}
\label{table:granularity-policy}
\vspace{-0.7cm}
\end{table}

\begin{table}[t]
\caption{Results of removing skip-policy in Eclipse. 
Uniform-$n$ means uniformly sampling $n$ frames from the video for reasoning. 
}
\vspace{-0.1cm}
 \setlength{\tabcolsep}{2pt}
\scriptsize
\begin{center}
\setlength\abovecaptionskip{-2cm}   
\setlength\belowcaptionskip{-2cm} 
\begin{tabular}{c|ccc|c}
\hline
Models & ~~Uniform-10~~ &~~ Uniform-20~~ & ~~Uniform-40~~ & ~Eclipse~(skip-policy) \\\hline
Accuracy & 34.16 & 35.49 & 36.48 & 37.05 \\ \hline
GFLOPs & 32.17 & 51.19 & 68.41 & 28.14 \\ \hline
\end{tabular}
\end{center}
\label{table:skipping-policy}
\vspace{-0.7cm}
\end{table}

\begin{table}[tbp]
\caption{Results of removing exit-policy in Eclipse. 
Final-Step Inference refers to that we remove the exit-policy and infer until the final step. 
}
\vspace{-0.1cm}
 \setlength{\tabcolsep}{10pt}
\scriptsize
\begin{center}
\begin{tabular}{c|c|c}
\hline
Models & Final-Step Inference & ~Eclipse~(exit-policy)\\\hline
Accuracy & 37.11  & 37.05 \\ \hline
GFLOPs & 36.92 & 28.14 \\ \hline
\end{tabular}
\end{center}
\label{table:exit-policy}
\vspace{-0.9cm}
\end{table}

\begin{figure}[htbp]
\vspace{-0.3cm}
\begin{center}
   \includegraphics[width=0.8\linewidth]{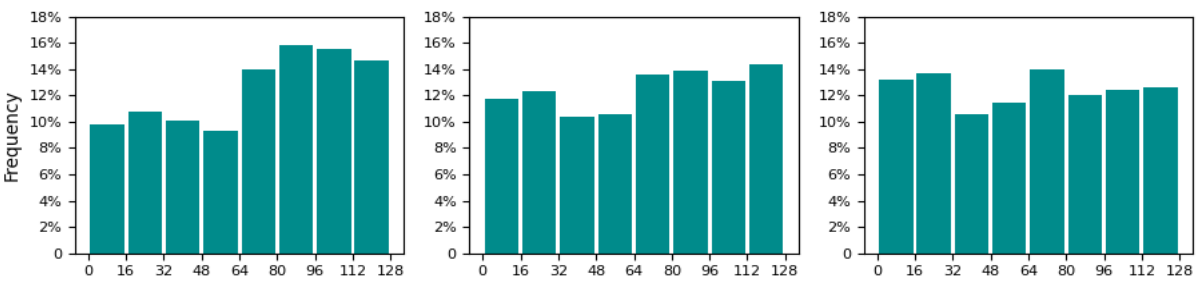}
\end{center}
\vspace{-0.4cm}
   \caption{Distributions of frame location (Left to right:step1, 2, 3).}
\label{fig:Freq}
\vspace{-0.4cm}
\end{figure}

{\bf Ablation Study.} As shown in Table \ref{table:granularity-policy}, compared with the model variant of using fine features only, by adopting the granularity-policy to adaptively decide feature granularity for the selected frame at each step, our network achieves nearly the same accuracy yet using much lower computation cost. 
Our network also achieves obviously higher reasoning accuracy than the method of using coarse features only. These results show the effectiveness of our granularity-policy by choosing fine features for most useful frames and coarse features for less important frames. 
Furthermore, we remove the skip-policy and simply uniformly sample frames at each step. 
As shown in Table \ref{table:skipping-policy}, our Eclipse performs the best yet at small computation cost. 
This shows that our skip-policy effectively reduces the computation cost by selecting useful frames for dynamic reasoning.
Moreover, we present the frame location distributions for the first three inference steps of our network in Figure \ref{fig:Freq}. 
As shown, our network selects frames dynamically for different videos as we expect. 
We also investigate the exit-policy module by comparing it with the method of reasoning until the final inference step. The result in Table \ref{table:exit-policy} shows that our model achieves best accuracy-to-computation ratio via adaptive inference. In Figure \ref{fig:policy_example}, we further present a qualitative example from our dataset to show how Eclipse performs dynamic and efficient reasoning. 

\section{Conclusion}
We create a new video QA dataset, SUTD-TrafficQA, focusing on video reasoning over traffic events. In our dataset, we introduce 6 reasoning tasks requiring various levels of causal reasoning. Besides, we propose the Eclipse network for video QA. By learning the dynamic glimpse policy and adaptive exit policy, our network achieves superior performance with significant computation efficiency.

{\bf Acknowledgement.} We would like to thank Yutian Lin, Renhang Liu, Yingjie Qiao, Xun Long Ng, Tran Nguyen Bao Long, Koh Kai Ting and Christabel Dorothy for their help in dataset collection and running baseline models. This work is supported by SUTD Projects PIE-SGP-Al2020-02 and SRG-ISTD-2020-153.

{\small
\bibliographystyle{ieee_fullname}
\bibliography{trafficqa_arxiv}
}

\end{document}